\documentclass[final,5p,times,twocolumn]{elsarticle}

\usepackage{dsfont}
\usepackage{setspace}

\usepackage[table]{xcolor}

\usepackage{caption}

\usepackage{pifont}
\usepackage{algorithm}
\usepackage{algpseudocode}
\usepackage{graphicx}

\usepackage{subfigure}
\usepackage{amsmath}

\usepackage{amssymb}
\usepackage{xcolor}
\usepackage{tabularx} 
\usepackage{multirow}
\usepackage{pbox}
\usepackage{amssymb}
\usepackage{amsfonts}
\usepackage{bbm}
\usepackage{graphicx}
{\begin{list}               
    {$\bullet$ \hfill}{
        \setlength{\leftmargin}{\parindent}
        \setlength{\parsep}{0.04\baselineskip}
        \setlength{\itemsep}{0.5\parsep}
        \setlength{\labelwidth}{\leftmargin}
        \setlength{\labelsep}{0em}}
    }
{\end{list}}

\providecommand{\cref}[1]{Chapter~\ref{#1}}

\providecommand{\fref}[1]{Figure~\ref{#1}}
\providecommand{\tref}[1]{Table~\ref{#1}}

\providecommand{\calD}{\mathcal{D}}

\providecommand{\calL}{\mathcal{L}}

\providecommand{\calY}{\mathcal{Y}}

\journal{Neurocomputing}

\begin{document}

\begin{frontmatter}


\title{ Long-Tailed Learning for Generalized Category Discovery}

\author[1]{Cuong Manh Hoang\corref{cor1}}
\ead{cuonghoang@seoultech.ac.kr}

\cortext[cor1]{Corresponding author.}
\affiliation[1]{organization={Department of Electronic Engineering, Seoul National University of Science and Technology},
            addressline={232 Gongneung-ro, Nowon-gu}, 
            city={Seoul},
            postcode={01811}, 
            country={South Korea}}

\begin{abstract}
Generalized Category Discovery (GCD) utilizes labeled samples of known classes to discover novel classes in unlabeled samples. Existing methods show effective performance on artificial datasets with balanced distributions. However, real-world datasets are always imbalanced, significantly affecting the effectiveness of these methods. To solve this problem, we propose a novel framework that performs generalized category discovery in long-tailed distributions. We first present a self-guided labeling technique that uses a learnable distribution to generate pseudo-labels, resulting in less biased classifiers. We then introduce a representation balancing process to derive discriminative representations. By mining sample neighborhoods, this process encourages the model to focus more on tail classes. We conduct experiments on public datasets to demonstrate the effectiveness of the proposed framework. The results show that our model exceeds previous state-of-the-art methods.
\end{abstract}


\end{frontmatter}



\section{INTRODUCTION}

Generalized Category Discovery (GCD) has recently attracted a lot of attention. By only leveraging available annotations, GCD models can classify known classes while jointly discovering unknown classes in the unlabeled data. This technique shows its practicality by saving a lot of effort in annotation. In real-world scenarios, this task is even more challenging because class distributions are typically imbalanced, with a few dominant classes and many others represented by only a small number of samples \cite{liu2019large,zhang2023deep}. Long-tailed distributions also arise in fields such as classification \cite{van2018inaturalist}, healthcare \cite{marrakchi2021fighting}, and autonomous driving \cite{mao2021one}.

Recently, methods for generalized category discovery (GCD) can effectively discover unknown classes by leveraging annotations for only the known classes. However, most of these methods assume a similar number of samples per class. The first approach \cite{vaze2022generalized} uses contrastive representation learning and semi-supervised k-means for discovering unseen classes. However, contrastive learning is vulnerable to the long-tailed data \cite{jiang2021self}. Moreover, k-means tends to focus on dense, majority regions of the data, leading to poor cluster quality for rare classes. SimGCD \cite{wen2023parametric} and ProtoGCD \cite{10948388} propose pseudo-labeling methods with entropy maximization regularizers \cite{hu2017learning,assran2022masked}, which lead to majority class samples being misclassified as minority class instances.

In the context of long-tailed generalized category discovery, BaCon \cite{bai2023towards} achieves state-of-the-art results by proposing a self-balanced co-advice contrastive framework. Specifically, they use a contrastive-learning branch to provide reliable distribution estimation that regularizes the predictions of the pseudo-labeling branch. The outputs from the pseudo-labeling branch then guide contrastive learning through self-balanced knowledge transfer and a proposed novel contrastive loss. Despite obtaining impressive results, BaCon \cite{bai2023towards} struggles to effectively handle several issues. Firstly, they estimate class frequencies by $k$-means, which generates poor results in imbalanced scenarios. The estimated distribution then regularizes the mean predictions of the pseudo-labeling branch with a limited batch size, which fails to capture a wide variety of classes in long-tailed distributions and results in noisy cluster assignments. Secondly, their contrastive-learning branch mainly depends on the unsupervised contrastive loss, which is biased towards head classes and performs poorly on tail classes.

\begin{figure*}[!t] \begin{center}
\begin{minipage}{0.48\linewidth}
\centerline{\includegraphics[scale=1.5]{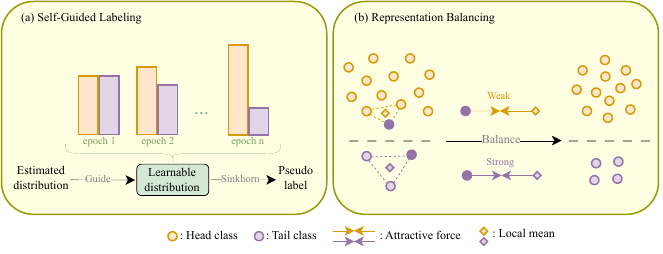}}
\end{minipage}

  \caption{(a) Self-guided labeling incorporates a learnable distribution that accurately reflects a long-tailed distribution during training; (b) Representation balancing leverages sample neighborhoods to direct the focus of the model on tail classes, deriving discriminative representations.}
\label{fig:teaser}

\end{center} \end{figure*}

To address these limitations, this paper introduces a novel framework that effectively performs long-tailed learning for generalized category discovery. Our model consists of a backbone, followed by a classification head and a projection head. The training process is divided into two stages: generalized category discovery and representation balancing. In the first stage, we follow GCD \cite{vaze2022generalized} to perform contrastive representation learning with the projection head, while applying a new self-guided labeling process to train with the classification head. Specifically, we introduce a learnable data distribution, which is guided by the estimated class distribution. Due to the long-tailed nature of the data and the absence of annotations, we apply long-tailed clustering on the extracted features to compute class sizes, which are then used to derive the estimated class distribution. The learnable distribution is subsequently combined with the Sinkhorn-Knopp algorithm \cite{cuturi2013sinkhorn} to produce pseudo-labels. To mitigate the problems caused by the limited batch size in long-tailed learning, we perform pseudo-labeling with a large queue of data samples \cite{he2020momentum}, which includes a diverse range of classes. While self-guided labeling can mitigate biased classifiers in long-tailed learning, contrastive representation learning remains vulnerable to imbalanced distributions \cite{jiang2021self}, which degrades the model performance. To address this issue, we propose a second training phase, called representation balancing. During this stage, only the backbone and the projection head, initialized from the previous stage, are used for training. By measuring the neighborhood density of a sample in the learned representation space, we can approximate its likelihood of belonging to head or tail classes. For instance, samples from tail classes typically exhibit lower neighborhood density. This helps the model identify samples from tail classes and focus on them, resulting in more discriminative representations. Figure 1 provides a brief overview of our main contributions. We evaluate the proposed framework on public datasets. The experimental results show that our proposed method outperforms the previous state-of-the-art methods.

The contributions of this paper are summarized as follows: (1) We propose a novel framework that effectively performs long-tailed learning for generalized category discovery; (2) We present a new self-guided labeling method to avoid biased classifiers. To generate high-quality pseudo-labels aligned with the actual data distribution, we employ a learnable target distribution guided by an estimated distribution derived through long-tailed clustering; (3) We introduce a new representation balancing technique to produce more discriminative representations. By calculating the neighborhood density of samples in the learned representation space, the model can identify tail class samples and focus more on them; (4) We validate the effectiveness of the proposed framework by freezing the optimized backbone and applying clustering to its extracted features. Experiments are conducted on public datasets.

\section{RELATED WORKS}
\subsection{Long-tailed Supervised Learning}

Long-tailed learning has received significant attention in recent years because of the difficulties associated with highly imbalanced datasets. This distribution makes models easily biased towards dominant classes and leads to poor performance on tail classes. In supervised learning, there are many approaches to address this problem.

Re-sampling is one of the earliest strategies for long-tailed supervised learning. SMOTE \cite{chawla2002smote} introduces an over-sampling technique by repeating data from the rare classes, while EUCI \cite{liu2008exploratory} proposes an under-sampling by discarding a portion of data from the frequent classes. However, studies \cite{zhou2020bbn, cao2019learning} point out that over-sampling often results in overfitting on the tail classes, and under-sampling can remove valuable information which inevitably degrades the model performance. Recently, DLC \cite{kang2019decoupling} introduces class-balanced re-sampling where samples from each class have an identical probability of being sampled. This method can mitigate problems caused by previous models.
 
Another common approach is re-weighting techniques, which adjust the training loss values for different classes by multiplying them with different weights. The simplest
method is to apply label frequencies from the training samples for loss re-weighting. Class-balanced loss (CB) \cite{cui2019class} introduces a concept of the effective number to approximate the expected sample number of different classes, which is an exponential function of their number of training labels. Then, they apply a class-balanced re-weighting factor, which is inversely proportional to the effective number of classes, to mitigate class imbalance. Instead of using training label frequencies, Equalization loss (EQL) \cite{tan2020equalization} directly down-weights the loss values of tail-class samples when they serve as negative labels for head-class samples. However, simply down-weighting negative gradients may harm the discriminative abilities of deep models. To address this, Adaptive Class Suppression loss (ACSL) \cite{wang2021adaptive} uses the output confidence to determine whether to suppress the gradient for a negative label. 

Apart from the aforementioned approaches, data augmentation also attracts significant attention. M2m \cite{kim2020m2m} introduces a transfer-based augmentation method that augments tail classes by translating head-class samples into tail-class ones via perturbation-based optimization. MISLAS \cite{zhong2021improving} presents a non-transfer augmentation technique by using data mixup to enhance representation learning in a decoupled scheme. Recently, CMO \cite{park2022majority} combines tail instances with head class images used as backgrounds to enhance tail class diversity. However, this approach overlooks semantic similarity by mixing images at random. OTmix \cite{gao2023enhancing} addresses this issue through an adaptive image-mixing strategy grounded in optimal transport (OT), which integrates both class-level and instance-level information. Data augmentation can also be conducted in the feature space. For example, H2T \cite{li2024feature} enriches tail class diversity by fusing feature maps from head class samples with those of the tail classes.

Rather than synthesizing new images for minority classes, GCL \cite{li2022long} proposes a Gaussian clouded logit adjustment approach to make tail class samples more active. Later, UCLD \cite{chen2023transfer} introduces a knowledge transfer-based calibration method that computes and applies importance weights to samples from tail classes. Another method, SHIKE \cite{jin2023long}, introduces a self-heterogeneous integration framework that aggregates knowledge from multiple experts and facilitates knowledge transfer among them.

With pre-trained foundation models, LPT \cite{dong2023lpt} introduces a prompt tuning method to tackle long-tailed learning. To achieve better generalization on both head and tail classes, GNM-PT \cite{li2024improving} fine-tunes pre-trained models using an innovative Gaussian Neighborhood Minimization optimizer. LIFT \cite{shi2024long} identifies the negative impact of heavy fine-tuning on tail-class performance and proposes a more efficient fine-tuning approach.

Although delivering impressive performance, these techniques require labels for all classes, which are expensive to collect. To address this issue, we propose a novel framework for long-tailed generalized category discovery that utilizes available annotations to classify known classes while also discovering unknown classes in the unlabeled data. In our setting, the distributions of both known classes and unknown classes are long-tailed.

\subsection{Generalized / Novel Class Discovery}
Given labeled and unlabeled datasets, novel class discovery (NCD) aims to discover unknown classes in the unlabeled dataset that differ from the known classes in the labeled dataset. It uses knowledge from the labeled dataset to discover and cluster unseen objects into distinct semantic categories. There are two main groups according to their clustering methods. The first group typically explores pairwise similarity for clustering. CCN \cite{hsu2018learning} suggests leveraging predictive pairwise similarity as the knowledge to be transferred and formulates a learnable objective function that uses this pairwise information in a manner similar to constrained clustering. Autonovel \cite{han2021autonovel} introduces robust rank statistics to evaluate the similarity of two data points in their representation space. The second group presents self-labeling for clustering unknown classes. UNO \cite{fini2021unified} uses the Sinkhorn-Knopp algorithm \cite{cuturi2013sinkhorn} to generate pseudo-labels, which are treated homogeneously with ground truth labels, enabling a single CE loss to operate on both labeled and unlabeled sets. ComEx \cite{yang2022divide} interprets pseudo-labeling as a global-to-local alignment between cluster centers and training samples, and proposes to strengthen pseudo labels with local-to-local aggregation among neighborhood samples. Although these methods show significant progress, they assume that unseen classes are uniformly distributed. For long-tailed distributions, NCDLR \cite{zhangnovel} introduces a class representation method based on equiangular prototypes alongside an iterative pseudo-label generation strategy for visual class learning. They specifically model the pseudo-label generation process as a relaxed optimal transport problem and implement a bi-level optimization algorithm to solve it efficiently.

While NCD makes the unrealistic assumption that all of the unlabeled images come from new categories, generalized category discovery (GCD) presents a realistic vision setting where the unlabelled images come from both known and unknown categories. The first approach for GCD \cite{vaze2022generalized} initially learns the representations using unsupervised and supervised contrastive losses. Then, the learned features are clustered by semi-supervised $k$-means for label assignment. To optimize performance and minimize computational costs, SimGCD \cite{wen2023parametric} proposes to use a classification head along with the representation head. Besides a supervised classification loss, an unsupervised classification loss with a mean-entropy maximisation regulariser is also used to train the classification head. For more balanced accuracy between old and new classes, ProtoGCD \cite{10948388} proposes a unified and unbiased prototype learning framework that models both old and new classes using shared prototypes and common learning objectives, facilitating consistent representation across all classes. To address long-tailed distributions, BaCon \cite{bai2023towards} designs a novel self-balanced co-advice contrastive framework. This framework comprises a contrastive-learning branch and a pseudo-labeling branch that work together to address the data imbalance and classify unseen categories simultaneously. However, they use classical $k$-means to estimate class frequencies, which regularize the mean prediction of the pseudo-labeling branch within the limited batch size, leading to poor performance in long-tailed distributions. Moreover, their contrastive-learning branch mainly depends on the unsupervised contrastive loss, which is not suitable for long-tailed distributions.

To overcome these limitations, we first propose a new self-guided labeling process to avoid non-activated classifiers. Specifically, we apply an efficient long-tailed clustering technique to estimate class frequencies, which are used to guide a learnable distribution. To generate high-quality pseudo-labels, the Sinkhorn-Knopp algorithm \cite{cuturi2013sinkhorn} is then applied with the learnable distribution to a large queue of data samples \cite{he2020momentum}, which contains a wide variety of classes. Furthermore, we present a new representation balancing technique to generate discriminative representations by leveraging learned representations of sample neighborhoods.

\section{PROPOSED METHOD}
\label{sec:method}
\begin{figure*}[!t] \begin{center}
\begin{minipage}{0.49\linewidth}
\centerline{\includegraphics[scale=1.5]{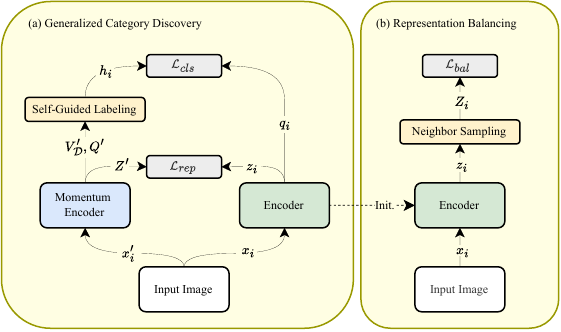}}
\end{minipage}
   \caption{Illustration of the proposed framework. (a) Generalized category discovery includes a self-guided labeling process to produce pseudo-labels ($h_i$) for the classification loss ($\calL_{cls}$). (b) Representation balancing computes the balanced loss ($\calL_{bal}$) based on the sample neighborhood ($Z_i$).}
\label{fig:overview}
\end{center}\end{figure*}

In this section, we first define the problem in our settings. Subsequently, we describe the generalized category discovery, introducing the self-guided labeling technique. Finally, we present the representation balancing process. The overview of the proposed framework is shown in~\fref{fig:overview}. 

\subsection{Problem Formulation}
\label{sec:pf}
In our setting, the training set $\calD$ comprises the labeled set $\mathcal{D}_{l}=\{(x_i,y_i)\}_{i=1}^{M_1} \in \mathcal{X} \times \mathcal{Y}_l $ and the unlabeled set $\mathcal{D}_{u}=\{(x_i,y_i)\}_{i=1}^{M_2} \in \mathcal{X} \times \mathcal{Y}_u $. $\calY_n=\{\calY_u \setminus \calY_l\}$ denotes novel classes and $M=M_1+M_2$ represents the total number of images in the training set. We do not have access to the labels of $\calD_u$ during training. The class distributions of $\calD_l$ and $\calD_u$ are both long-tailed with imbalance ratios $\rho_l, \rho_u \gg 1$. After training, the model must assign labels to a disjoint balanced test set including all classes of $\calY_u$. Following BaCon \cite{bai2023towards}, we assume that the number of classes in $\calD$ is known in advance.    

\subsection{Generalized Category Discovery}
\label{sec:gcd}
In long-tailed distributions, a limited batch size may fail to capture a wide variety of classes, degrading the performance \cite{fomenko2022learning}. For this reason, we propose our model based on MoCo \cite{he2020momentum}. Given an image $x_i$, a backbone $b(.)$ is used to extract its features $v_i=b(x_i)$. Then, a classification head $c(.)$ and a projection head $p(.)$ are used to generate the probability $q_i=c(v_i) \in \mathbb{R}^{1\times |\calY_u|}$ and the representation $z_i=p(v_i) \in \mathbb{R}^{1\times d}$, where $d$ is the feature dimension. Following MoCo \cite{he2020momentum}, this encoder is used to update a momentum encoder, which includes $\bar{b}(.)$, $\bar{c}(.)$, and $\bar{p}(.)$. With another view of the image $x_i'$, the momentum encoder generates $v_i'=\bar{b}(x_i')$, $q_i'=\bar{c}(v_i')$, and $z_i'=\bar{p}(v_i')$. During training, the queues $Q'=\{q_i'\}_{i=1}^N \in \mathbb{R}^{N\times |\calY_u|}$ and $Z'=\{z_i'\}_{i=1}^N \in \mathbb{R}^{N\times d}$, where $N$ is the queue size, are used to store the probabilities $q_i'$ and the representations $z_i'$ of samples from both previous and current mini-batches. These large queues contain a diverse range of classes used to compute classification and representation objectives.

For the classification objective, a new self-guided labeling technique is proposed to avoid biased classifiers. We first estimate the class distribution using a long-tailed clustering technique. 
Given the extracted features from the momentum backbone $V'_{\calD}=\{v_i'\}_{i=1}^M$, we initialize $C$ prototypes $\{\hat{v}_c\}_{c=1}^C$ using $k$-means, with $C=|\mathcal{Y}_u|$. Then, the probability $\phi_{ic}$ that sample $x_i'$ belongs to cluster $c$ is calculated as follows:
\begin{equation}
	\phi_{ic}=\frac{\text{exp}(\text{sim}(v'_i,\hat{v}_c))}{\sum_{c'=1}^C\text{exp}(\text{sim}(v'_i,\hat{v}_{c'}))}
 \label{eq:propa}
\end{equation}
where $\text{sim}(.)$ is the cosine similarity. Next, we compute the target assignment probability as follows:
\begin{equation}
\begin{split}
	&\Phi_{ic}=\frac{\text{exp}(\alpha\text{sim}(v'_i,\hat{v}_c))}{\sum_{c'=1}^C\text{exp}(\alpha\text{sim}(v'_i,\hat{v}_{c'}))}\\
    &\text{where} \; \alpha=\gamma-\frac{1}{M}\sum_{i=1}^M\text{sim}(v'_i,\hat{v}_{c})
 \label{eq:target}
 \end{split}
\end{equation}
where $\gamma$ is a hyperparameter. With $\alpha>1$, the target assignment probability can promote learning from high confidence cluster assignments while mitigating issues related to imbalanced clusters. Subsequently, the prototypes are updated by the cluster loss as follows:
\begin{equation}
	\calL_{cluster}=\frac{1}{M}\sum_{i=1}^{M}\sum_{c=1}^C\Phi_{ic}\text{log}\frac{\Phi_{ic}}{\phi_{ic}}
 \label{eq:clusterloss}
\end{equation}
After the convergence, each sample $x_i'$ is assigned to their corresponding cluster by $\text{argmax}_{c\in \{1,2,...,C\}}\phi_{ic} $ to get sizes of clusters $\{\tilde{n}_c\}_{c=1}^{C}$. Following BaCon \cite{bai2023towards}, these sizes are used to obtain the class distribution $\tilde{\pi}=\{\tilde{\pi}_c\}_{c=1}^{C}$ with $\tilde{\pi}_c=\tilde{n}_c/\sum_{c=1}^C\tilde{n}_c$, and the distribution is then assigned to known and unknown categories. We estimate the class distribution every $T_1$ epochs for efficiency. 
   
To generate pseudo-labels, we apply the Sinkhorn-Knopp algorithm \cite{cuturi2013sinkhorn} to the queue $Q'$. However, directly using $\tilde{\pi}$ as the target distribution for the Sinkhorn-Knopp algorithm \cite{cuturi2013sinkhorn} is suboptimal, as clustering is an unsupervised technique that struggles to precisely capture the true underlying data distribution. To overcome this limitation, we introduce a learnable target distribution $\pi$ with sizes of clusters $\{n_c\}_{c=1}^C$ as parameters. With $\pi$, the Sinkhorn-Knopp algorithm \cite{cuturi2013sinkhorn} is applied to $Q'$ to generate pseudo-labels $H(\pi,Q')=\{h_i\}_{i=1}^N \in \mathbb{R}^{N\times |\calY_u|}$. To ensure high-quality pseudo-labels, we compute a guided loss as follows:
\begin{equation}
	\calL_{gud}=-\frac{1}{N}\sum_{i=1}^{N}\sum_{k=1}^{|\calY{u}|}q_{ik}'h_{ik} + \beta\text{KL}(\pi,\tilde{\pi})
 \label{eq:constraint}
\end{equation}
where $\beta$ is a hyperparameter. With this loss, $\pi$ can learn to align with both the clustering-based estimate $\tilde{\pi}$ and the distribution observed in the queue $Q'=\{q_i'\}_{i=1}^N$ , resulting in a more accurate representation of the real data distribution. Next, the generated pseudo-labels are used to train the encoder with an unsupervised classification loss as follows:
\begin{equation}
	\calL_{cls}^u=-\frac{1}{B}\sum_{i=1}^{B}\sum_{k=1}^{|\calY{u}|}h_{ik}\text{log}q_{ik} 
 \label{eq:ucross}
\end{equation}
where $B$ is the batch size. Apart from the unsupervised cross-entropy loss, we also compute supervised cross-entropy loss $\calL_{cls}^s$ with available labels. Then, the classification objective is calculated as follows:
\begin{equation}
	\calL_{cls}=(1-\lambda)\calL_{cls}^u+\lambda\calL_{cls}^s+\calL_{gud}
 \label{eq:cls}
\end{equation}
where $\lambda$ is a hyperparameter.

For the representation objective, we follow GCD \cite{vaze2022generalized} to compute the unsupervised contrastive loss $\calL_{rep}^u$ and the supervised contrastive loss $\calL_{rep}^s$. In these losses, negative samples are taken from the queue $Z'$ to capture a wide variety of classes. Then, the representation objective is computed as follows:
\begin{equation}
\calL_{rep}=(1-\lambda)\calL_{rep}^u+\lambda\calL_{rep}^s
\label{eqn:repob}
\end{equation}
Finally, the overall objective is $\calL_{cls}+\calL_{rep}$.
\subsection{Representation Balancing}
\label{sec:rb}
In the previous stage, the representation objective mainly depends on the unsupervised contrastive loss. However, SDCLR \cite{jiang2021self} demonstrated that this loss is vulnerable to class imbalance, which may struggle to generalize to long-tailed data challenges. To resolve this problem, we propose a new balancing process that leverages sample neighborhoods to produce discriminative representations in long-tailed distributions.

Given the optimized encoder from the previous stage, we use its backbone $b(.)$ and its projection head $p(.)$ to initialize the encoder in this stage. With an input image $x_i$, we extract its representation $z_i$. Then, we sample a set $Z_i$, which contains $z_i$ and its $K$ nearest neighbors in the representations of the training set. This sampling process is implemented every $T_2$ epochs. 

Using $Z_i$, we estimate local density by computing the cosine similarity between all pairs of the samples in $Z_i$ as follows:
\begin{equation}
	w_{i}=-\frac{\sum_{m \in Z_i}\sum_{n \in Z_i}\mathds{1}_{[n \neq m]}\text{sim}(z_m,z_n)}{K(K+1)}
 \label{eq:dens}
\end{equation}
Since samples belonging to head classes tend to have higher densities in the embedding space than those of tail classes, the value of $w_{i}$ is low when $x_i$ belongs to head classes and high when $x_i$ belongs to tail classes. Next, we compute the local mean of $Z_i$ as follows:
\begin{equation}
	\mu_{i}=\frac{1}{K+1}\sum_{m \in Z_i}z_m
 \label{eq:mean}
\end{equation}
The local mean $\mu_i$ represents the neighborhood of $z_i$ and contains more semantic information than $z_i$. With $w_i$ and $\mu_i$, we calculate a balanced loss as follows:
\begin{equation}
	\calL_{bal}=\frac{1}{B}\sum_{i=1}^B(1+w_i)(1-\text{sim}(z_i,\mu_i))
 \label{eq:blloss}
\end{equation}
With this loss function, the model becomes less biased toward head classes by assigning lower values of $w_i$ to head class samples, and focuses more on tail classes by assigning higher values of $w_i$ to tail class samples. Furthermore, by maximizing the cosine similarities between samples and their local means, the model can encode the semantic structure of the data while producing discriminative representations. This loss function is computed entirely in an unsupervised manner.  

After training, the optimized backbone is used for inference to extract features, which are clustered to generate predictions.

\section{EXPERIMENTS AND RESULTS}
\label{sec:result}

\subsection{Experimental setting}
\noindent \textbf{Dataset.} Following BaCon \cite{bai2023towards}, we perform experiments with four public datasets, including CIFAR-10-LT, CIFAR-100-LT, ImageNet-100-LT, and Places-LT. CIFAR-10-LT/CIFAR-100-LT are sampled from CIFAR10/CIFAR100 \cite{cui2019class} with exponential distribution. ImageNet-100-LT is sampled from ImageNet-100 \cite{tian2020contrastive} with Pareto distribution. Places-365-LT is sampled from Places-365 \cite{zhou2017places} with Pareto distribution. For all datasets, $|\calY_l|$ classes from the long-tailed datasets are used as known classes, while the remaining classes $|\calY_n|$ are treated as novel classes. Half of the instances in each known class are used as $\calD_l$, while the remaining instances are used to form $\calD_u$. In the default setting, $\calY_l$ for CIFAR-10-LT, CIFAR-100-LT, ImageNet-100-LT, and Places-365-LT is set to 5, 80, 50, and 182, respectively. Additionally, the imbalance ratios $\rho=\rho_l=\rho_u$ for these datasets are 100, 100, 256, and 996, respectively. The summary of these datasets are shown in \tref{tab:data}.

\begin{table*}[!ht]
\small
\begin{center}
\begin{minipage}{0.9\linewidth}
\caption{Summary of long-tailed datasets used in our default setting.}
\label{tab:data}
\begin{tabular}{>{\raggedright}m{0.2\textwidth}|>{\centering}m{0.168\textwidth}>{\centering}m{0.168\textwidth}>{\centering\arraybackslash}m{0.168\textwidth}>{\centering\arraybackslash}m{0.168\textwidth}} 
 \hline
Characteristics & CIFAR-10-LT&CIFAR-100-LT& ImageNet-100-LT& Places-365-LT\\ 
\hline\hline
Long-tail type & Exp & Exp & Pareto&Pareto \\ 
Imbalance ratio $(\rho)$ & 100 & 100 & 256 & 996 \\ 
$\#$ known classes $(|\calY_l|)$ & 5 & 80 & 50 &182 \\ 
$\#$ novel classes $(|\calY_n|)$& 5 & 20 & 50 &183 \\
$\#$ labeled images $(M_1)$& 4.5K & 3.8K & 3.2K&20.8K \\ 
$\#$ unlabeled images $(M_2)$& 6.7K & 6.0K & 9.0K&41.7K \\ 
$\#$ test images & 10.0K & 10.0K & 5.0K &36.5K \\ 
 
\hline
\end{tabular}
\end{minipage}
\end{center}
\end{table*}

\begin{table*}[!t]
\small
\begin{center}
\begin{minipage}{0.9\linewidth}
\caption{ Test accuracy (\%) on four long-tailed datasets. (*: adapted methods, \textbf{bold}: the best score, \underline{underline}: the second-best score.)}
\label{tab:result_all}
\begin{tabular}{>{\raggedright}m{0.12\textwidth}|>{\centering}m{0.045\textwidth}>{\centering}m{0.045\textwidth}>{\centering\arraybackslash}m{0.045\textwidth}|>{\centering\arraybackslash}m{0.045\textwidth}>{\centering\arraybackslash}m{0.045\textwidth}>{\centering\arraybackslash}m{0.045\textwidth}|>{\centering\arraybackslash}m{0.045\textwidth}>{\centering\arraybackslash}m{0.045\textwidth}>{\centering\arraybackslash}m{0.045\textwidth}|>{\centering\arraybackslash}m{0.045\textwidth}>{\centering\arraybackslash}m{0.045\textwidth}>{\centering\arraybackslash}m{0.045\textwidth}}  
\hline
\multirow{2}{*}{Methods} & \multicolumn{3}{c|}{CIFAR-10-LT} & \multicolumn{3}{c|}{CIFAR-100-LT}& \multicolumn{3}{c|}{ImageNet-100-LT}& \multicolumn{3}{c}{Places-365-LT} \\
\cline{2-13}
 & Old & New & All &  Old & New & All & Old & New & All& Old & New & All\\
\hline\hline
$\text{ABC}^*$ \cite{lee2021abc}& 77.7 & 20.2 & 48.9 & 48.5 &20.8& 43.0& - & - & - & -& - & -  \\ 
$\text{DARP}^*$ \cite{kim2020distribution}&  77.6 &35.2 &56.4& 54.5& 24.0& 48.4 & - & - & - & -& - & -  \\ 
$\text{TRSSL}^*$ \cite{rizve2022towards}& 78.4& 66.8& 72.6& 58.7 &35.8 &54.1& - & - & - & - & - & -   \\ 
$\text{ORCA}^*$ \cite{cao2022openworld}& 77.5 &55.6& 66.6& 55.0& 30.8& 50.1& 69.3& 29.0& 49.2& 21.5& 6.9& 14.2 \\ 
SimGCD \cite{wen2023parametric}& 75.1& 41.5& 58.3& 59.8& 24.6 &52.8 &81.1 &33.4 &57.8& \underline{31.4}& 18.2 &24.8  \\ 
GCD \cite{vaze2022generalized}& 78.5& 71.7 &75.1 &65.5& 49.0 &62.2& 81.0 &76.8 &78.9 &29.8& 22.7 &26.2  \\ 
ProtoGCD \cite{10948388}& 84.3& 74.9 &79.6 &65.9& 51.7 &63.1& 81.9 &74.5&78.2 &30.4& 24.6 &27.5  \\ 
$\text{OpenCon}^*$ \cite{sun2023opencon}& 87.2& 47.2& 67.2& 64.2& 40.9 &59.6 &83.3& 42.8 &63.0 &30.6& 12.4& 21.6  \\ 
$\text{NCDLR}^*$ \cite{zhangnovel}& 89.4 & 81.2 & 85.3 & 65.8 & 55.2 & 63.7 & 83.7 & 79.2 & 81.5 & 30.8 & 25.7 & 28.2  \\ 
BaCon \cite{bai2023towards}& \underline{94.2}& \underline{88.1}& \underline{91.1}& \underline{67.4}& \underline{66.5}& \underline{67.2}& \underline{84.6}& \underline{82.8}& \underline{83.7} &31.1& \underline{28.4} &\underline{29.9}  \\ 
Ours& \textbf{96.7} & \textbf{93.3} & \textbf{95.0} & \textbf{74.5}& \textbf{73.8} &\textbf{74.4} & \textbf{89.0} & \textbf{87.8} & \textbf{88.4} & \textbf{33.2}& \textbf{31.7}& \textbf{32.5} \\ 

 \hline
\end{tabular}
\end{minipage}
\end{center}
\end{table*}

\noindent \textbf{Implementation Details.} In generalized category discovery, we use a pre-trained ViT-B/16 \cite{caron2021emerging} model to initialize the backbone. The classification head and the projection head are randomly initialized. The batch size $B$ and the queue size $N$ are set to 256 and 2048, respectively. The hyperparameter $\gamma=2$, $\beta=400$, $\lambda=0.35$, $T_1=10$. For representation balancing, we set $K=5$ and $T_2=10$. The experiments were conducted on a computer with two Nvidia GeForce RTX 3090 GPUs, an Intel Core i9-10940X CPU, and 128 GB RAM.

\noindent \textbf{Evaluation Metric.} Following BaCon \cite{bai2023towards}, we use the Hungarian assignment algorithm to map the predictions to the ground truth. After that, the results are evaluated on $\calY_l$ (Old), $\calY_n$ (New), and $\calY_l \cup \calY_n$ (All). Additionally, we divide the test dataset into three disjoint groups (i.e. Many, Medium, Few) according to the number of instances per class. We then assess the balancedness by calculating the standard deviation of the accuracies for these three groups.
\subsection{Result}

\begin{table*}[!t]
\small
\begin{center}
\begin{minipage}{0.9\linewidth}
\caption{Test accuracy (\%) and balancedness (Std$\downarrow$) on CIFAR-100-LT \cite{cui2019class}.}
\label{tab:result_balance}
\begin{tabular}{>{\raggedright}m{0.12\textwidth}|>{\centering}m{0.05\textwidth}>{\centering}m{0.05\textwidth}>{\centering\arraybackslash}m{0.05\textwidth}>{\centering\arraybackslash}m{0.05\textwidth}>{\centering\arraybackslash}m{0.05\textwidth}|>{\centering\arraybackslash}m{0.05\textwidth}>{\centering\arraybackslash}m{0.05\textwidth}>{\centering\arraybackslash}m{0.05\textwidth}>{\centering\arraybackslash}m{0.05\textwidth}>{\centering\arraybackslash}m{0.05\textwidth}|>{\centering\arraybackslash}m{0.06\textwidth}}  
\hline
\multirow{2}{*}{Methods} & \multicolumn{5}{c|}{Known} & \multicolumn{5}{c|}{Novel} & Overall \\
\cline{2-11}
 & Many$\uparrow$ & Med$\uparrow$ & Few$\uparrow$ & Std$\downarrow$ & All$\uparrow$ & Many$\uparrow$ & Med$\uparrow$ & Few$\uparrow$ & Std$\downarrow$ & All$\uparrow$ & \\
\hline\hline
$\text{ABC}^*$ \cite{lee2021abc}& 77.4& 51.6& 16.3& 25.0 &48.5 &20.8 &32.0& 5.9 &10.7& 20.8 &43.0\\
$\text{DARP}^*$ \cite{kim2020distribution}& 74.8& 55.2& 33.4 &16.9& 54.5& 29.9& 30.1& 10.0& 9.4& 24.0 &48.4\\
$\text{TRSSL}^*$ \cite{rizve2022towards}& 78.8& \underline{73.0}& 24.1 &24.3& 58.7& 32.7& 50.9& 18.8 &13.1& 35.8& 54.1\\
$\text{ORCA}^*$ \cite{cao2022openworld} & 73.6 & 60.0 & 31.0 & 17.8 & 55.0 & 47.5 & 28.4 & 17.2 & 12.5 & 30.8 & 50.1 \\ 
SimGCD \cite{wen2023parametric} & 71.1 & 72.8 & 34.6 & 17.6 & 59.8 & 32.9 & 25.8 & 15.5 & 7.1 & 24.6 & 52.8 \\ 
GCD \cite{vaze2022generalized} & 75.9 & 69.4 & 52.9 & 9.7 & 65.5 & 41.9 & 54.2 & 49.8 & \underline{5.1} & 49.0 & 62.2 \\ 
ProtoGCD \cite{10948388}& 76.2& 67.8 &53.7 &9.3& 65.9 &57.5& 52.4 &45.2 &5.2&51.7& 63.1 \\ 
$\text{OpenCon}^*$ \cite{sun2023opencon} & \textbf{86.8} & 69.7 & 35.8 & 21.2 & 64.2 & 55.8 & 48.9 & 15.3 & 17.7 & 40.9 & 59.6 \\ 
$\text{NCDLR}^*$ \cite{zhangnovel} & 75.7 & 65.5 & 55.8 & 8.8 & 65.8 & 61.7 & 53.2 & 50.8 & 5.4 & 55.2 & 63.7 \\ 
BaCon \cite{bai2023towards} & 74.4 & 67.1 & \underline{60.8} & \underline{5.6} & \underline{67.4} & \underline{73.7} & \underline{66.1} & \underline{59.8} & 5.7 & \underline{66.5} & \underline{67.2} \\ 
Ours & \underline{80.9} & \textbf{73.6} & \textbf{69.2} & \textbf{4.8} & \textbf{74.5} & \textbf{79.6} & \textbf{73.1} & \textbf{68.8} & \textbf{4.5} & \textbf{73.8} & \textbf{74.4} \\ 
\hline
\end{tabular}
\end{minipage}
\end{center}
\end{table*}

\begin{table*}[!t]
\small
\begin{center}
\begin{minipage}{0.9\linewidth}
\caption{ Test accuracy (\%) on CIFAR-100-LT \cite{cui2019class} with different imbalance ratio $\rho$.}
\label{tab:result_rho}
\begin{tabular}{>{\raggedright}m{0.12\textwidth}|>{\centering}m{0.045\textwidth}>{\centering}m{0.045\textwidth}>{\centering}m{0.045\textwidth}|>{\centering}m{0.045\textwidth}>{\centering}m{0.045\textwidth}>{\centering}m{0.045\textwidth}|>{\centering}m{0.045\textwidth}>{\centering}m{0.045\textwidth}>{\centering}m{0.045\textwidth}|>{\centering}m{0.045\textwidth}>{\centering}m{0.045\textwidth}>{\centering\arraybackslash}m{0.045\textwidth}}  
\hline
\multirow{2}{*}{Methods} & \multicolumn{3}{c|}{$\rho=20$} & \multicolumn{3}{c|}{$\rho=50$}& \multicolumn{3}{c|}{$\rho=100$}& \multicolumn{3}{c}{$\rho=150$} \\
\cline{2-13}
 & Old & New & All &  Old & New & All & Old & New & All& Old & New & All\\
\hline\hline
$\text{ABC}^*$ \cite{lee2021abc}& 64.0& 23.0& 55.8& 53.7& 21.5& 47.3& 48.5& 20.8& 43.0& 41.9 &18.5& 37.2 \\ 
$\text{DARP}^*$ \cite{kim2020distribution}& 66.8&27.4& 58.9& 58.9& 25.1 &52.1& 54.5 &24.0 &48.4& 49.7& 23.3& 44.3  \\ 
$\text{TRSSL}^*$ \cite{rizve2022towards}& 71.8& 43.6& 66.2& 62.9& 49.0& 60.1& 58.7 &35.8 &54.1& 55.2& 33.3& 50.8 \\ 
$\text{ORCA}^*$ \cite{cao2022openworld}& 66.9& 37.7& 61.0& 61.7& 30.0& 55.3 &55.0& 30.8 &50.1& 51.0& 36.5& 48.1  \\ 
SimGCD \cite{wen2023parametric}&\underline{71.8} &23.1& 62.1& 63.7& 23.4& 55.7& 59.8& 24.2& 52.8 &57.5 &19.9 &50.0 \\ 
GCD \cite{vaze2022generalized}&  69.8& 58.7& 67.6& 66.6& 59.7& 65.2 &65.5 &49.0 &62.2 &65.4 &50.2& 62.4 \\ 
ProtoGCD \cite{10948388}& 71.2& 58.9 &68.7 &68.3& 56.8 &66.0& 65.9& 51.7&63.1 &64.7& 50.8& 61.9  \\ 
$\text{OpenCon}^*$ \cite{sun2023opencon}& 76.3& 44.3 &69.9 &\underline{71.2} &42.0& 65.4& 64.2& 40.9 &59.6 &59.7& 45.3& 56.8 \\ 
$\text{NCDLR}^*$ \cite{zhangnovel}& 73.2 & 59.2 & 70.4 & 69.8 & 54.3 & 66.7 & 65.8 & 55.2 & 63.7 & 63.4 & 52.9 & 61.3  \\ 
BaCon \cite{bai2023towards}& 71.6& \underline{68.5}& \underline{71.0} &70.4& \underline{64.2}& \underline{69.2}& \underline{67.4}& \underline{66.5}& \underline{67.2} &\underline{67.0} &\underline{64.3}& \underline{66.5}  \\ 
Ours& \textbf{78.1} & \textbf{76.6} & \textbf{77.8} & \textbf{76.7}& \textbf{74.2} &\textbf{76.2} & \textbf{74.5} & \textbf{73.8} & \textbf{74.4} & \textbf{73.9}& \textbf{72.5}& \textbf{73.6} \\

 \hline
\end{tabular}
\end{minipage}
\end{center}
\end{table*}

To demonstrate the effectiveness, our model is compared with existing frameworks, including ABC \cite{lee2021abc}, DARP \cite{kim2020distribution}, TRSSL \cite{rizve2022towards}, ORCA \cite{cao2022openworld}, SimGCD \cite{wen2023parametric}, GCD \cite{vaze2022generalized}, ProtoGCD \cite{10948388}, OpenCon \cite{sun2023opencon}, NCDLR \cite{zhangnovel}, and BaCon \cite{bai2023towards}. 

\tref{tab:result_all} shows quantitative results on CIFAR-10-LT \cite{cui2019class}, CIFAR-100-LT \cite{cui2019class}, ImageNet-100-LT \cite{tian2020contrastive}, and Places-365-LT \cite{zhou2017places}. In the table, we use boldface and underlines to denote the best and the second-best scores, respectively. * denotes adapted methods. The results show that our model surpasses the state-of-the-art BaCon \cite{bai2023towards} by 3.9$\%$, 7.2$\%$, 4.7$\%$, and 2.6$\%$ in overall accuracy for CIFAR-10-LT, CIFAR-100-LT, ImageNet-100-LT, and Places-365-LT, respectively. This superiority indicates the effectiveness of our contributions.

\tref{tab:result_balance} shows a detailed comparison on CIFAR-100-LT \cite{cui2019class}. The table shows that our model achieves the best results across most metrics. For known classes, our model ranks second in accuracy for head classes (Many) with 80.9$\%$, and ranks first for the Median and Few groups, achieving 73.6$\%$ and 69.2$\%$, respectively. For novel classes, our model attains the highest accuracy across all groups, with 79.6$\%$, 73.1$\%$, and 68.8$\%$ for Many, Median, and Few, respectively. Compared to the state-of-the-art BaCon \cite{bai2023towards}, our model significantly outperforms it by 8.4$\%$ and 9.0$\%$ on the known and novel minority classes (Few), respectively. The table also shows that our model produces the most balanced features, with standard deviations (Std) of 4.8 and 4.5 for known classes and novel classes, respectively. It proves our effectiveness in balancing long-tailed distributions.
 
\tref{tab:result_rho} presents the results under varying imbalance ratios $\rho = \rho_l = \rho_u$ on the CIFAR-100-LT dataset \cite{cui2019class}. Our model consistently achieves the highest performance across all settings. With imbalance ratios of $\rho = 20$, $50$, and $100$, our model attains the highest overall accuracies of 77.8$\%$, 76.2$\%$, and 74.4$\%$, respectively. Even in the highly imbalanced scenario with $\rho = 150$, it still achieves the best overall accuracy of 73.6$\%$, outperforming the state-of-the-art BaCon \cite{bai2023towards} by 7.1$\%$. These results demonstrate that our model remains effective across different imbalance ratios.

\begin{figure*}[t] 
\begin{center}

\begin{minipage}{0.3\linewidth}
\centerline{\includegraphics[scale=0.27]{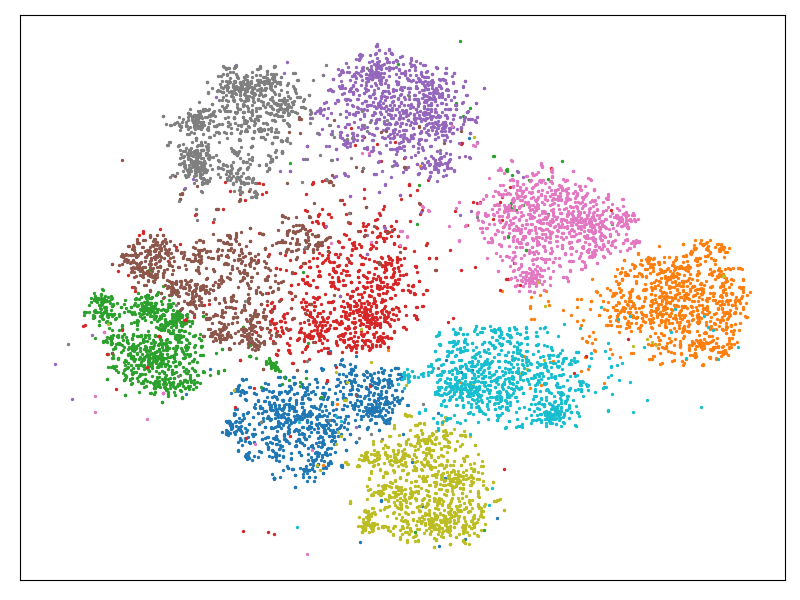}}
\end{minipage}
\begin{minipage}{0.3\linewidth}
\centerline{\includegraphics[scale=0.27]{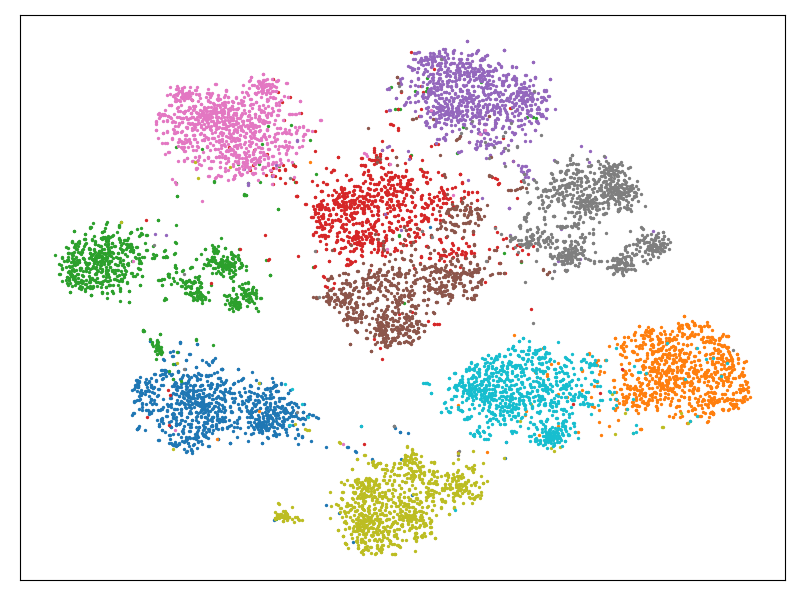}}
\end{minipage}
\begin{minipage}{0.3\linewidth}
\centerline{\includegraphics[scale=0.27]{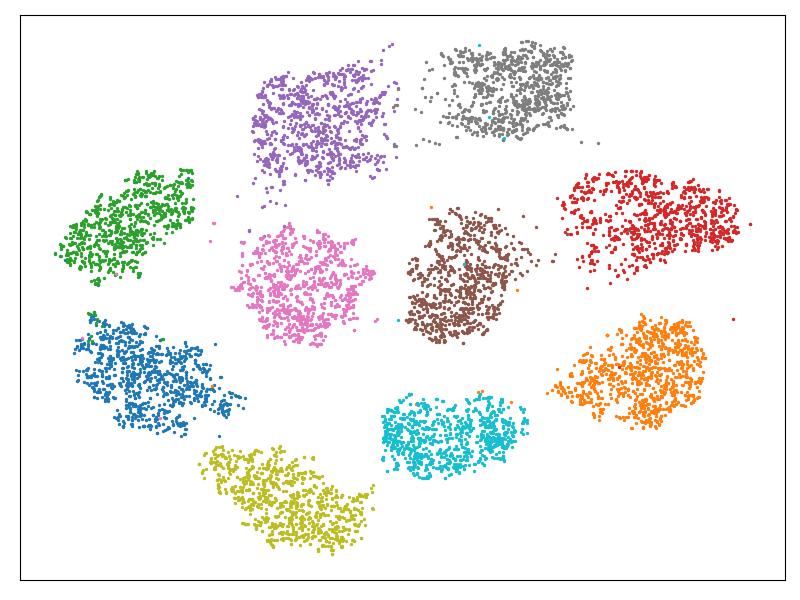}}
\end{minipage}
\\

\vspace{0.1cm}

\begin{minipage}{0.3\linewidth}
\centerline{\small (a) GCD \cite{vaze2022generalized}}
\end{minipage}
\begin{minipage}{0.3\linewidth}
\centerline{\small (b) Bacon \cite{bai2023towards}}
\end{minipage}
\begin{minipage}{0.3\linewidth}
\centerline{\small (c) Our}
\end{minipage}

\caption{t-SNE visualization on the test set of CIFAR-10 \cite{cui2019class}. }
\label{fig:tsne}
\end{center}
\end{figure*}

\fref{fig:tsne} visualizes a t-SNE projection of the learned representation on the test set of CIFAR-10 \cite{cui2019class}. This figure verifies that the proposed method achieves superior separability between samples of different classes and creates more compact clusters for each class compared to previous approaches.

\tref{tab:subtab1} shows an ablation study on the components of the proposed framework using CIFAR-100-LT \cite{cui2019class}. The baseline model represents the framework without self-guided labeling (SGL) and representation balancing (RB). For the classification objective in the baseline, the Sinkhorn-Knopp algorithm \cite{cuturi2013sinkhorn} is applied to achieve balanced class distributions. The table shows that by using the self-guided labeling (SGL) to avoid non-activated classifiers in long-tailed distributions, the overall accuracy increases by 8.4$\%$. Then, incorporating representation balancing (RB) to focus more on tail classes leads to an increase of 7.6$\%$ in overall accuracy. Moreover, our model significantly surpasses the baseline by 40.1$\%$ in accuracy for novel classes (New). This indicates the effectiveness of our model in discovering unknown classes with long-tailed distributions.

\tref{tab:subtab2} shows an ablation study of different classifier training techniques using CIFAR-100-LT \cite{cui2019class}. Firstly, when using regularized predictions (RP) from BaCon \cite{bai2023towards}, our model achieves an overall accuracy of 69.8$\%$. Then, the overall accuracy increases by 3.4$\%$, when we apply our self-guided labeling with $k$-means (Ours-$k$). This demonstrates the effectiveness of our labeling technique using a learnable distribution. Finally, by using our long-tailed clustering technique, the overall accuracy increases by 1.2$\%$. This is because $k$-means typically positions most cluster centroids near head class samples, with only a few near tail class samples, while our clustering technique effectively resolves this issue.

\begin{table*}[t]
\centering
\small
\captionof{table}{Ablation studies on the components of the proposed framework using CIFAR-100-LT \cite{cui2019class}.}
\label{tab:compo}
\captionsetup[subtable]{skip=2pt}

\begin{minipage}[t]{0.29\textwidth}
    \centering
    \captionof{subtable}{Contribution of each component.} 
    \label{tab:subtab1}
    \begin{tabular}{l|ccc}
    \hline
    Methods & Old & New & All \\
    \hline
    \hline
    Baseline & 64.6 & 33.7&58.4\\
    +SGL& 70.0 &53.9 &66.8\\
    +RB& \textbf{74.5} &\textbf{73.8} &\textbf{74.4}\\
    \hline
    \end{tabular}
\end{minipage} 
\begin{minipage}[t]{0.3\textwidth}
    \centering
    \captionof{subtable}{ Different methods for classifier training.} 
    \label{tab:subtab2}
    \begin{tabular}{l|ccc}
    \hline
    Methods & Old & New & All \\
    \hline
    \hline
    RP \cite{bai2023towards} & 70.1 &68.7&69.8\\
    Ours-$k$ & 73.6 &71.5&73.2\\
    Ours & \textbf{74.5} &\textbf{73.8} &\textbf{74.4}\\

    \hline
    \end{tabular}
\end{minipage} 
\\
\vspace{0.5cm}
\begin{minipage}[t]{0.33\textwidth}
    \centering
    \captionof{subtable}{Target distributions for pseudo-labeling.}
    \label{tab:subtab3}
    \begin{tabular}{l|ccc}
    \hline
    Distributions & Old & New & All \\
    \hline
    \hline
    Uniform  & 66.9 & 58.7 & 65.3 \\
    Estimated & 72.0 & 67.5 & 71.1 \\
    Learnable(Ours) & \textbf{74.5} & \textbf{73.8} & \textbf{74.4} \\
    \hline
    \end{tabular}
\end{minipage}
\begin{minipage}[t]{0.35\textwidth}
    \centering
    \captionof{subtable}{Different methods for representation learning.} 
    \label{tab:subtab4}
    \begin{tabular}{l|ccc}
    \hline
    Methods & Old & New & All \\
    \hline
    \hline
    SDCLR \cite{jiang2021self}& 69.1&67.3&68.7\\
    TS \cite{kukleva2023temperature} & 71.2 &69.7&70.9\\
    CONMIX \cite{li2025conmix} & 73.7 & 71.1 & 73.2 \\
    Ours & \textbf{74.5} &\textbf{73.8} &\textbf{74.4}\\

    \hline
    \end{tabular}
\end{minipage}

\end{table*}

\begin{table*}[t!]
\centering
\small
\captionof{table}{Ablation studies on the hyperparameters using CIFAR-100-LT \cite{cui2019class}. Our default settings are marked in \colorbox{gray!30}{gray}.}
\label{tab:hyper}
\captionsetup[subtable]{skip=2pt}

\begin{minipage}[t]{0.23\textwidth}
    \centering
    \captionof{subtable}{$T_1$ for distribution estimation} 
    \label{tab:subtable1}
    \begin{tabular}{c|ccc}
    \hline
    $T_1$ & Old & New & All \\
    \hline
    \hline
    1 & 74.7 & \textbf{74.5}&\textbf{74.7}\\
    5& \textbf{74.8} &73.1 &74.5\\
    \cellcolor{gray!30}10&  \cellcolor{gray!30}74.5   &\cellcolor{gray!30}73.8 & \cellcolor{gray!30}74.4\\
    25 & 74.6&70.9 &73.9 \\
    50 & 74.3&70.0 &73.5 \\
    \hline
    \end{tabular}
\end{minipage} 
\begin{minipage}[t]{0.23\textwidth}
    \centering
    \captionof{subtable}{$\beta$ for guided loss} 
    \label{tab:subtable2}
    \begin{tabular}{c|ccc}
    \hline
    $\beta$ & Old & New & All \\
    \hline
    \hline
    200 & 74.2 &71.9&73.8\\
    300 & 74.4 &72.7&74.1\\
    \cellcolor{gray!30}400&\cellcolor{gray!30}\textbf{74.5}  &\cellcolor{gray!30}\textbf{73.8} & \cellcolor{gray!30}\textbf{74.4} \\
    500 & 74.3 &73.6&74.2\\
    600 & 74.1 &73.3&74.0\\
    \hline
    \end{tabular}
\end{minipage} 
\begin{minipage}[t]{0.23\textwidth}
    \centering
    \captionof{subtable}{$K$ for neighbor sampling} 
    \label{tab:subtable3}
    \begin{tabular}{c|ccc}
    \hline
    $K$ & Old & New & All \\
    \hline
    \hline
    2 & 73.8&70.6 &73.2\\
    3 & 74.3 &71.6&73.8\\
    \cellcolor{gray!30}5&  \cellcolor{gray!30}74.5   &\cellcolor{gray!30}73.8 & \cellcolor{gray!30}74.4\\
    6 &74.4&\textbf{74.7} &74.5 \\
    7 &\textbf{74.7} &74.1&\textbf{74.6} \\
    \hline
    \end{tabular}
\end{minipage} 
\begin{minipage}[t]{0.23\textwidth}
    \centering
    \captionof{subtable}{$T_2$ for neighbor sampling} 
    \label{tab:subtable4}
    \begin{tabular}{c|ccc}
    \hline
    $T_2$ & Old & New & All \\
    \hline
    \hline
    1 & \textbf{74.7} &\textbf{74.0}&\textbf{74.6}\\
    5 & 74.6 &73.3&74.4\\
    \cellcolor{gray!30}10&  \cellcolor{gray!30}74.5   &\cellcolor{gray!30}73.8 & \cellcolor{gray!30}74.4\\
    25 & 74.2 &73.0&74.0\\
    50 & 73.8 &71.3&73.3\\
    \hline
    \end{tabular}
\end{minipage}

\end{table*}

\tref{tab:subtab3} presents an ablation study on different target distributions for the Sinkhorn-Knopp algorithm \cite{cuturi2013sinkhorn} using CIFAR-100-LT \cite{cui2019class}. The results indicate that the estimated distribution obtained through long-tailed clustering boosts overall accuracy by 5.8$\%$ compared to the uniform distribution. Furthermore, employing the learnable target distribution results in an additional 3.3$\%$ improvement in overall accuracy, indicating that it more effectively captures the true underlying data distribution.

\tref{tab:subtab4} shows an ablation study of different representation learning techniques using CIFAR-100-LT \cite{cui2019class}. For comparison, we train our model using only the first stage and compute the unsupervised
contrastive loss for long-tailed distributions as SDCLR \cite{jiang2021self}, TS \cite{kukleva2023temperature}, and CONMIX \cite{li2025conmix}. By leveraging sample neighborhoods to focus more on tail classes, our representation balancing outperforms SDCLR \cite{jiang2021self}, TS \cite{kukleva2023temperature} and CONMIX \cite{li2025conmix} by 5.7$\%$ 3.5$\%$, and 1.2$\%$ in overall accuracy, respectively.

In \tref{tab:hyper}, we present ablation studies of the hyperparameters using CIFAR-100-LT \cite{cui2019class}. \tref{tab:subtable1} presents the performance with different values of $T_1$. It can be observed that the overall accuracy increases by estimating the distribution more frequently. In \tref{tab:subtable2}, the proposed method consistently delivers good performance, even when suboptimal values of $\beta$ are employed. In \tref{tab:subtable3}, increasing the number of sampled neighbors $K$ improves accuracy by capturing more local information and semantic representations. Finally, \tref{tab:subtable4} shows that reducing the interval for neighbor sampling results in better performance. In our default setting, the values of the hyperparameters are selected to ensure not only the effectiveness but also the efficiency of the model.

\section{CONCLUSION}
In this paper, we present a novel framework for generalized category discovery in long-tailed distributions. To avoid non-activated classifiers, a self-guided labeling technique with a learnable distribution is first proposed to produce high-quality pseudo-labels. To generate discriminative representations, a representation balancing process is then introduced to help the model focus more on tail classes by mining sample neighborhoods. The experimental results indicate that the proposed method outperforms previous state-of-the-art methods on public datasets.

A limitation of the proposed model is that it requires labels of known classes during training. In the future, we plan to design a framework that can effectively perform clustering in long-tailed distributions without the need for human annotations.  



\section*{Acknowledgments}
The author would like to express sincere thanks to the Editor-in-Chief Professor Zidong Wang, the Editors and anonymous reviewers for their valuable comments and suggestions, which greatly improved this paper.

\bibliographystyle{elsarticle-harv}
\bibliography{2_2_mybibfile}

\begin{thebibliography}{49}
\expandafter\ifx\csname natexlab\endcsname\relax\def\natexlab#1{#1}\fi
\providecommand{\url}[1]{\texttt{#1}}
\providecommand{\href}[2]{#2}
\providecommand{\path}[1]{#1}
\providecommand{\DOIprefix}{doi:}
\providecommand{\ArXivprefix}{arXiv:}
\providecommand{\URLprefix}{URL: }
\providecommand{\Pubmedprefix}{pmid:}
\providecommand{\doi}[1]{\href{http://dx.doi.org/#1}{\path{#1}}}
\providecommand{\Pubmed}[1]{\href{pmid:#1}{\path{#1}}}
\providecommand{\bibinfo}[2]{#2}
\ifx\xfnm\relax \def\xfnm[#1]{\unskip,\space#1}\fi
\bibitem[{Assran et~al.(2022)Assran, Caron, Misra, Bojanowski, Bordes, Vincent,
  Joulin, Rabbat and Ballas}]{assran2022masked}
\bibinfo{author}{Assran, M.}, \bibinfo{author}{Caron, M.},
  \bibinfo{author}{Misra, I.}, \bibinfo{author}{Bojanowski, P.},
  \bibinfo{author}{Bordes, F.}, \bibinfo{author}{Vincent, P.},
  \bibinfo{author}{Joulin, A.}, \bibinfo{author}{Rabbat, M.},
  \bibinfo{author}{Ballas, N.}, \bibinfo{year}{2022}.
\newblock \bibinfo{title}{Masked siamese networks for label-efficient
  learning}, in: \bibinfo{booktitle}{European conference on computer vision},
  \bibinfo{organization}{Springer}. pp. \bibinfo{pages}{456--473}.
\bibitem[{Bai et~al.(2023)Bai, Liu, Wang, Chen, Mu, Li, Zhou, Feng, Wu and
  Hu}]{bai2023towards}
\bibinfo{author}{Bai, J.}, \bibinfo{author}{Liu, Z.}, \bibinfo{author}{Wang,
  H.}, \bibinfo{author}{Chen, R.}, \bibinfo{author}{Mu, L.},
  \bibinfo{author}{Li, X.}, \bibinfo{author}{Zhou, J.T.},
  \bibinfo{author}{Feng, Y.}, \bibinfo{author}{Wu, J.}, \bibinfo{author}{Hu,
  H.}, \bibinfo{year}{2023}.
\newblock \bibinfo{title}{Towards distribution-agnostic generalized category
  discovery}.
\newblock \bibinfo{journal}{Advances in Neural Information Processing Systems}
  \bibinfo{volume}{36}, \bibinfo{pages}{58625--58647}.
\bibitem[{Cao et~al.(2022)Cao, Brbic and Leskovec}]{cao2022openworld}
\bibinfo{author}{Cao, K.}, \bibinfo{author}{Brbic, M.},
  \bibinfo{author}{Leskovec, J.}, \bibinfo{year}{2022}.
\newblock \bibinfo{title}{Open-world semi-supervised learning}, in:
  \bibinfo{booktitle}{International Conference on Learning Representations},
  pp. \bibinfo{pages}{512--534}.
\newblock \URLprefix \url{https://openreview.net/forum?id=O-r8LOR-CCA}.
\bibitem[{Cao et~al.(2019)Cao, Wei, Gaidon, Arechiga and Ma}]{cao2019learning}
\bibinfo{author}{Cao, K.}, \bibinfo{author}{Wei, C.}, \bibinfo{author}{Gaidon,
  A.}, \bibinfo{author}{Arechiga, N.}, \bibinfo{author}{Ma, T.},
  \bibinfo{year}{2019}.
\newblock \bibinfo{title}{Learning imbalanced datasets with
  label-distribution-aware margin loss}.
\newblock \bibinfo{journal}{Advances in neural information processing systems}
  \bibinfo{volume}{32}.
\bibitem[{Caron et~al.(2021)Caron, Touvron, Misra, J{\'e}gou, Mairal,
  Bojanowski and Joulin}]{caron2021emerging}
\bibinfo{author}{Caron, M.}, \bibinfo{author}{Touvron, H.},
  \bibinfo{author}{Misra, I.}, \bibinfo{author}{J{\'e}gou, H.},
  \bibinfo{author}{Mairal, J.}, \bibinfo{author}{Bojanowski, P.},
  \bibinfo{author}{Joulin, A.}, \bibinfo{year}{2021}.
\newblock \bibinfo{title}{Emerging properties in self-supervised vision
  transformers}, in: \bibinfo{booktitle}{Proceedings of the IEEE/CVF
  international conference on computer vision}, pp.
  \bibinfo{pages}{9650--9660}.
\bibitem[{Chawla et~al.(2002)Chawla, Bowyer, Hall and
  Kegelmeyer}]{chawla2002smote}
\bibinfo{author}{Chawla, N.V.}, \bibinfo{author}{Bowyer, K.W.},
  \bibinfo{author}{Hall, L.O.}, \bibinfo{author}{Kegelmeyer, W.P.},
  \bibinfo{year}{2002}.
\newblock \bibinfo{title}{Smote: synthetic minority over-sampling technique}.
\newblock \bibinfo{journal}{Journal of artificial intelligence research}
  \bibinfo{volume}{16}, \bibinfo{pages}{321--357}.
\bibitem[{Chen and Su(2023)}]{chen2023transfer}
\bibinfo{author}{Chen, J.}, \bibinfo{author}{Su, B.}, \bibinfo{year}{2023}.
\newblock \bibinfo{title}{Transfer knowledge from head to tail: Uncertainty
  calibration under long-tailed distribution}, in:
  \bibinfo{booktitle}{Proceedings of the IEEE/CVF conference on computer vision
  and pattern recognition}, pp. \bibinfo{pages}{19978--19987}.
\bibitem[{Cui et~al.(2019)Cui, Jia, Lin, Song and Belongie}]{cui2019class}
\bibinfo{author}{Cui, Y.}, \bibinfo{author}{Jia, M.}, \bibinfo{author}{Lin,
  T.Y.}, \bibinfo{author}{Song, Y.}, \bibinfo{author}{Belongie, S.},
  \bibinfo{year}{2019}.
\newblock \bibinfo{title}{Class-balanced loss based on effective number of
  samples}, in: \bibinfo{booktitle}{Proceedings of the IEEE/CVF conference on
  computer vision and pattern recognition}, pp. \bibinfo{pages}{9268--9277}.
\bibitem[{Cuturi(2013)}]{cuturi2013sinkhorn}
\bibinfo{author}{Cuturi, M.}, \bibinfo{year}{2013}.
\newblock \bibinfo{title}{Sinkhorn distances: Lightspeed computation of optimal
  transport}.
\newblock \bibinfo{journal}{Advances in neural information processing systems}
  \bibinfo{volume}{26}.
\bibitem[{Dong et~al.(2023)Dong, Zhou, YAN and Zuo}]{dong2023lpt}
\bibinfo{author}{Dong, B.}, \bibinfo{author}{Zhou, P.}, \bibinfo{author}{YAN,
  S.}, \bibinfo{author}{Zuo, W.}, \bibinfo{year}{2023}.
\newblock \bibinfo{title}{{LPT}: Long-tailed prompt tuning for image
  classification}, in: \bibinfo{booktitle}{The Eleventh International
  Conference on Learning Representations}, pp. \bibinfo{pages}{41229--41248}.
\newblock \URLprefix \url{https://openreview.net/forum?id=8pOVAeo8ie}.
\bibitem[{Fini et~al.(2021)Fini, Sangineto, Lathuili{\`e}re, Zhong, Nabi and
  Ricci}]{fini2021unified}
\bibinfo{author}{Fini, E.}, \bibinfo{author}{Sangineto, E.},
  \bibinfo{author}{Lathuili{\`e}re, S.}, \bibinfo{author}{Zhong, Z.},
  \bibinfo{author}{Nabi, M.}, \bibinfo{author}{Ricci, E.},
  \bibinfo{year}{2021}.
\newblock \bibinfo{title}{A unified objective for novel class discovery}, in:
  \bibinfo{booktitle}{Proceedings of the IEEE/CVF International Conference on
  Computer Vision}, pp. \bibinfo{pages}{9284--9292}.
\bibitem[{Fomenko et~al.(2022)Fomenko, Elezi, Ramanan, Leal-Taix{\'e} and
  Osep}]{fomenko2022learning}
\bibinfo{author}{Fomenko, V.}, \bibinfo{author}{Elezi, I.},
  \bibinfo{author}{Ramanan, D.}, \bibinfo{author}{Leal-Taix{\'e}, L.},
  \bibinfo{author}{Osep, A.}, \bibinfo{year}{2022}.
\newblock \bibinfo{title}{Learning to discover and detect objects}.
\newblock \bibinfo{journal}{Advances in Neural Information Processing Systems}
  \bibinfo{volume}{35}, \bibinfo{pages}{8746--8759}.
\bibitem[{Gao et~al.(2023)Gao, Zhao, Li and Guo}]{gao2023enhancing}
\bibinfo{author}{Gao, J.}, \bibinfo{author}{Zhao, H.}, \bibinfo{author}{Li,
  Z.}, \bibinfo{author}{Guo, D.}, \bibinfo{year}{2023}.
\newblock \bibinfo{title}{Enhancing minority classes by mixing: An adaptative
  optimal transport approach for long-tailed classification}.
\newblock \bibinfo{journal}{Advances in Neural Information Processing Systems}
  \bibinfo{volume}{36}, \bibinfo{pages}{60329--60348}.
\bibitem[{Han et~al.(2021)Han, Rebuffi, Ehrhardt, Vedaldi and
  Zisserman}]{han2021autonovel}
\bibinfo{author}{Han, K.}, \bibinfo{author}{Rebuffi, S.A.},
  \bibinfo{author}{Ehrhardt, S.}, \bibinfo{author}{Vedaldi, A.},
  \bibinfo{author}{Zisserman, A.}, \bibinfo{year}{2021}.
\newblock \bibinfo{title}{Autonovel: Automatically discovering and learning
  novel visual categories}.
\newblock \bibinfo{journal}{IEEE Transactions on Pattern Analysis and Machine
  Intelligence} \bibinfo{volume}{44}, \bibinfo{pages}{6767--6781}.
\bibitem[{He et~al.(2020)He, Fan, Wu, Xie and Girshick}]{he2020momentum}
\bibinfo{author}{He, K.}, \bibinfo{author}{Fan, H.}, \bibinfo{author}{Wu, Y.},
  \bibinfo{author}{Xie, S.}, \bibinfo{author}{Girshick, R.},
  \bibinfo{year}{2020}.
\newblock \bibinfo{title}{Momentum contrast for unsupervised visual
  representation learning}, in: \bibinfo{booktitle}{Proceedings of the IEEE/CVF
  conference on computer vision and pattern recognition}, pp.
  \bibinfo{pages}{9729--9738}.
\bibitem[{Hsu et~al.(2018)Hsu, Lv and Kira}]{hsu2018learning}
\bibinfo{author}{Hsu, Y.C.}, \bibinfo{author}{Lv, Z.}, \bibinfo{author}{Kira,
  Z.}, \bibinfo{year}{2018}.
\newblock \bibinfo{title}{Learning to cluster in order to transfer across
  domains and tasks}, in: \bibinfo{booktitle}{International Conference on
  Learning Representations}, pp. \bibinfo{pages}{2567--2581}.
\bibitem[{Hu et~al.(2017)Hu, Miyato, Tokui, Matsumoto and
  Sugiyama}]{hu2017learning}
\bibinfo{author}{Hu, W.}, \bibinfo{author}{Miyato, T.}, \bibinfo{author}{Tokui,
  S.}, \bibinfo{author}{Matsumoto, E.}, \bibinfo{author}{Sugiyama, M.},
  \bibinfo{year}{2017}.
\newblock \bibinfo{title}{Learning discrete representations via information
  maximizing self-augmented training}, in: \bibinfo{booktitle}{International
  conference on machine learning}, \bibinfo{organization}{PMLR}. pp.
  \bibinfo{pages}{1558--1567}.
\bibitem[{Jiang et~al.(2021)Jiang, Chen, Mortazavi and Wang}]{jiang2021self}
\bibinfo{author}{Jiang, Z.}, \bibinfo{author}{Chen, T.},
  \bibinfo{author}{Mortazavi, B.J.}, \bibinfo{author}{Wang, Z.},
  \bibinfo{year}{2021}.
\newblock \bibinfo{title}{Self-damaging contrastive learning}, in:
  \bibinfo{booktitle}{International Conference on Machine Learning},
  \bibinfo{organization}{PMLR}. pp. \bibinfo{pages}{4927--4939}.
\bibitem[{Jin et~al.(2023)Jin, Li, Lu, Cheung and Wang}]{jin2023long}
\bibinfo{author}{Jin, Y.}, \bibinfo{author}{Li, M.}, \bibinfo{author}{Lu, Y.},
  \bibinfo{author}{Cheung, Y.m.}, \bibinfo{author}{Wang, H.},
  \bibinfo{year}{2023}.
\newblock \bibinfo{title}{Long-tailed visual recognition via self-heterogeneous
  integration with knowledge excavation}, in: \bibinfo{booktitle}{Proceedings
  of the IEEE/CVF conference on computer vision and pattern recognition}, pp.
  \bibinfo{pages}{23695--23704}.
\bibitem[{Kang et~al.(2020)Kang, Xie, Rohrbach, Yan, Gordo, Feng and
  Kalantidis}]{kang2019decoupling}
\bibinfo{author}{Kang, B.}, \bibinfo{author}{Xie, S.},
  \bibinfo{author}{Rohrbach, M.}, \bibinfo{author}{Yan, Z.},
  \bibinfo{author}{Gordo, A.}, \bibinfo{author}{Feng, J.},
  \bibinfo{author}{Kalantidis, Y.}, \bibinfo{year}{2020}.
\newblock \bibinfo{title}{Decoupling representation and classifier for
  long-tailed recognition}, in: \bibinfo{booktitle}{Eighth International
  Conference on Learning Representations (ICLR)}, pp.
  \bibinfo{pages}{5319--5328}.
\bibitem[{Kim et~al.(2020a)Kim, Hur, Park, Yang, Hwang and
  Shin}]{kim2020distribution}
\bibinfo{author}{Kim, J.}, \bibinfo{author}{Hur, Y.}, \bibinfo{author}{Park,
  S.}, \bibinfo{author}{Yang, E.}, \bibinfo{author}{Hwang, S.J.},
  \bibinfo{author}{Shin, J.}, \bibinfo{year}{2020}a.
\newblock \bibinfo{title}{Distribution aligning refinery of pseudo-label for
  imbalanced semi-supervised learning}.
\newblock \bibinfo{journal}{Advances in neural information processing systems}
  \bibinfo{volume}{33}, \bibinfo{pages}{14567--14579}.
\bibitem[{Kim et~al.(2020b)Kim, Jeong and Shin}]{kim2020m2m}
\bibinfo{author}{Kim, J.}, \bibinfo{author}{Jeong, J.}, \bibinfo{author}{Shin,
  J.}, \bibinfo{year}{2020}b.
\newblock \bibinfo{title}{M2m: Imbalanced classification via major-to-minor
  translation}, in: \bibinfo{booktitle}{Proceedings of the IEEE/CVF conference
  on computer vision and pattern recognition}, pp.
  \bibinfo{pages}{13896--13905}.
\bibitem[{Kukleva et~al.(2023)Kukleva, Böhle, Schiele, Kuehne and
  Rupprecht}]{kukleva2023temperature}
\bibinfo{author}{Kukleva, A.}, \bibinfo{author}{Böhle, M.},
  \bibinfo{author}{Schiele, B.}, \bibinfo{author}{Kuehne, H.},
  \bibinfo{author}{Rupprecht, C.}, \bibinfo{year}{2023}.
\newblock \bibinfo{title}{Temperature schedules for self-supervised contrastive
  methods on long-tail data}, in: \bibinfo{booktitle}{ICLR}, pp.
  \bibinfo{pages}{1683--1692}.
\newblock \URLprefix \url{https://openreview.net/forum?id=ejHUr4nfHhD}.
\bibitem[{Lee et~al.(2021)Lee, Shin and Kim}]{lee2021abc}
\bibinfo{author}{Lee, H.}, \bibinfo{author}{Shin, S.}, \bibinfo{author}{Kim,
  H.}, \bibinfo{year}{2021}.
\newblock \bibinfo{title}{Abc: Auxiliary balanced classifier for
  class-imbalanced semi-supervised learning}.
\newblock \bibinfo{journal}{Advances in Neural Information Processing Systems}
  \bibinfo{volume}{34}, \bibinfo{pages}{7082--7094}.
\bibitem[{Li et~al.(2022)Li, Cheung and Lu}]{li2022long}
\bibinfo{author}{Li, M.}, \bibinfo{author}{Cheung, Y.m.}, \bibinfo{author}{Lu,
  Y.}, \bibinfo{year}{2022}.
\newblock \bibinfo{title}{Long-tailed visual recognition via gaussian clouded
  logit adjustment}, in: \bibinfo{booktitle}{Proceedings of the IEEE/CVF
  conference on computer vision and pattern recognition}, pp.
  \bibinfo{pages}{6929--6938}.
\bibitem[{Li et~al.(2024a)Li, Liu, Lu, Zhang, Cheung and
  Huang}]{li2024improving}
\bibinfo{author}{Li, M.}, \bibinfo{author}{Liu, Y.}, \bibinfo{author}{Lu, Y.},
  \bibinfo{author}{Zhang, Y.}, \bibinfo{author}{Cheung, Y.m.},
  \bibinfo{author}{Huang, H.}, \bibinfo{year}{2024}a.
\newblock \bibinfo{title}{Improving visual prompt tuning by gaussian
  neighborhood minimization for long-tailed visual recognition}.
\newblock \bibinfo{journal}{Advances in Neural Information Processing Systems}
  \bibinfo{volume}{37}, \bibinfo{pages}{103985--104009}.
\bibitem[{Li et~al.(2024b)Li, Zhikai, Lu, Lan, Cheung and
  Huang}]{li2024feature}
\bibinfo{author}{Li, M.}, \bibinfo{author}{Zhikai, H.}, \bibinfo{author}{Lu,
  Y.}, \bibinfo{author}{Lan, W.}, \bibinfo{author}{Cheung, Y.m.},
  \bibinfo{author}{Huang, H.}, \bibinfo{year}{2024}b.
\newblock \bibinfo{title}{Feature fusion from head to tail for long-tailed
  visual recognition}, in: \bibinfo{booktitle}{Proceedings of the AAAI
  conference on artificial intelligence}, pp. \bibinfo{pages}{13581--13589}.
\bibitem[{Li and Jia(2025)}]{li2025conmix}
\bibinfo{author}{Li, Z.}, \bibinfo{author}{Jia, Y.}, \bibinfo{year}{2025}.
\newblock \bibinfo{title}{Conmix: Contrastive mixup at representation level for
  long-tailed deep clustering}, in: \bibinfo{booktitle}{The Thirteenth
  International Conference on Learning Representations}, pp.
  \bibinfo{pages}{653--667}.
\newblock \URLprefix \url{https://openreview.net/forum?id=3lH8WT0fhu}.
\bibitem[{Liu et~al.(2008)Liu, Wu and Zhou}]{liu2008exploratory}
\bibinfo{author}{Liu, X.Y.}, \bibinfo{author}{Wu, J.}, \bibinfo{author}{Zhou,
  Z.H.}, \bibinfo{year}{2008}.
\newblock \bibinfo{title}{Exploratory undersampling for class-imbalance
  learning}.
\newblock \bibinfo{journal}{IEEE Transactions on Systems, Man, and Cybernetics,
  Part B (Cybernetics)} \bibinfo{volume}{39}, \bibinfo{pages}{539--550}.
\bibitem[{Liu et~al.(2019)Liu, Miao, Zhan, Wang, Gong and Yu}]{liu2019large}
\bibinfo{author}{Liu, Z.}, \bibinfo{author}{Miao, Z.}, \bibinfo{author}{Zhan,
  X.}, \bibinfo{author}{Wang, J.}, \bibinfo{author}{Gong, B.},
  \bibinfo{author}{Yu, S.X.}, \bibinfo{year}{2019}.
\newblock \bibinfo{title}{Large-scale long-tailed recognition in an open
  world}, in: \bibinfo{booktitle}{Proceedings of the IEEE/CVF conference on
  computer vision and pattern recognition}, pp. \bibinfo{pages}{2537--2546}.
\bibitem[{Ma et~al.(2025)Ma, Zhu, Zhang and Liu}]{10948388}
\bibinfo{author}{Ma, S.}, \bibinfo{author}{Zhu, F.}, \bibinfo{author}{Zhang,
  X.Y.}, \bibinfo{author}{Liu, C.L.}, \bibinfo{year}{2025}.
\newblock \bibinfo{title}{Protogcd: Unified and unbiased prototype learning for
  generalized category discovery}.
\newblock \bibinfo{journal}{IEEE Transactions on Pattern Analysis and Machine
  Intelligence} ,
  \bibinfo{pages}{1--17}\DOIprefix\doi{10.1109/TPAMI.2025.3557502}.
\bibitem[{Mao et~al.(2021)Mao, Niu, Jiang, Liang, Liang, Li, Ye, Zhang, Li, Yu
  et~al.}]{mao2021one}
\bibinfo{author}{Mao, J.}, \bibinfo{author}{Niu, M.}, \bibinfo{author}{Jiang,
  C.}, \bibinfo{author}{Liang, H.}, \bibinfo{author}{Liang, X.},
  \bibinfo{author}{Li, Y.}, \bibinfo{author}{Ye, C.}, \bibinfo{author}{Zhang,
  W.}, \bibinfo{author}{Li, Z.}, \bibinfo{author}{Yu, J.}, et~al.,
  \bibinfo{year}{2021}.
\newblock \bibinfo{title}{One million scenes for autonomous driving: Once
  dataset}.
\newblock \bibinfo{journal}{NeurIPS} .
\bibitem[{Marrakchi et~al.(2021)Marrakchi, Makansi and
  Brox}]{marrakchi2021fighting}
\bibinfo{author}{Marrakchi, Y.}, \bibinfo{author}{Makansi, O.},
  \bibinfo{author}{Brox, T.}, \bibinfo{year}{2021}.
\newblock \bibinfo{title}{Fighting class imbalance with contrastive learning},
  in: \bibinfo{booktitle}{International conference on medical image computing
  and computer-assisted intervention}, \bibinfo{organization}{Springer}. pp.
  \bibinfo{pages}{466--476}.
\bibitem[{Park et~al.(2022)Park, Hong, Heo, Yun and Choi}]{park2022majority}
\bibinfo{author}{Park, S.}, \bibinfo{author}{Hong, Y.}, \bibinfo{author}{Heo,
  B.}, \bibinfo{author}{Yun, S.}, \bibinfo{author}{Choi, J.Y.},
  \bibinfo{year}{2022}.
\newblock \bibinfo{title}{The majority can help the minority: Context-rich
  minority oversampling for long-tailed classification}, in:
  \bibinfo{booktitle}{Proceedings of the IEEE/CVF conference on computer vision
  and pattern recognition}, pp. \bibinfo{pages}{6887--6896}.
\bibitem[{Rizve et~al.(2022)Rizve, Kardan and Shah}]{rizve2022towards}
\bibinfo{author}{Rizve, M.N.}, \bibinfo{author}{Kardan, N.},
  \bibinfo{author}{Shah, M.}, \bibinfo{year}{2022}.
\newblock \bibinfo{title}{Towards realistic semi-supervised learning}, in:
  \bibinfo{booktitle}{European Conference on Computer Vision},
  \bibinfo{organization}{Springer}. pp. \bibinfo{pages}{437--455}.
\bibitem[{Shi et~al.(2024)Shi, Wei, Zhou, Shao, Han and Li}]{shi2024long}
\bibinfo{author}{Shi, J.X.}, \bibinfo{author}{Wei, T.}, \bibinfo{author}{Zhou,
  Z.}, \bibinfo{author}{Shao, J.J.}, \bibinfo{author}{Han, X.Y.},
  \bibinfo{author}{Li, Y.F.}, \bibinfo{year}{2024}.
\newblock \bibinfo{title}{Long-tail learning with foundation model: Heavy
  fine-tuning hurts}, in: \bibinfo{booktitle}{International Conference on
  Machine Learning}, \bibinfo{organization}{PMLR}. pp.
  \bibinfo{pages}{45014--45039}.
\bibitem[{Sun and Li(2023)}]{sun2023opencon}
\bibinfo{author}{Sun, Y.}, \bibinfo{author}{Li, Y.}, \bibinfo{year}{2023}.
\newblock \bibinfo{title}{Opencon: Open-world contrastive learning}, in:
  \bibinfo{booktitle}{Transactions on Machine Learning Research}, pp.
  \bibinfo{pages}{5182--5194}.
\newblock \URLprefix \url{https://openreview.net/forum?id=2wWJxtpFer}.
\bibitem[{Tan et~al.(2020)Tan, Wang, Li, Li, Ouyang, Yin and
  Yan}]{tan2020equalization}
\bibinfo{author}{Tan, J.}, \bibinfo{author}{Wang, C.}, \bibinfo{author}{Li,
  B.}, \bibinfo{author}{Li, Q.}, \bibinfo{author}{Ouyang, W.},
  \bibinfo{author}{Yin, C.}, \bibinfo{author}{Yan, J.}, \bibinfo{year}{2020}.
\newblock \bibinfo{title}{Equalization loss for long-tailed object
  recognition}, in: \bibinfo{booktitle}{Proceedings of the IEEE/CVF conference
  on computer vision and pattern recognition}, pp.
  \bibinfo{pages}{11662--11671}.
\bibitem[{Tian et~al.(2020)Tian, Krishnan and Isola}]{tian2020contrastive}
\bibinfo{author}{Tian, Y.}, \bibinfo{author}{Krishnan, D.},
  \bibinfo{author}{Isola, P.}, \bibinfo{year}{2020}.
\newblock \bibinfo{title}{Contrastive multiview coding}, in:
  \bibinfo{booktitle}{Computer Vision--ECCV 2020: 16th European Conference,
  Glasgow, UK, August 23--28, 2020, Proceedings, Part XI 16},
  \bibinfo{organization}{Springer}. pp. \bibinfo{pages}{776--794}.
\bibitem[{Van~Horn et~al.(2018)Van~Horn, Mac~Aodha, Song, Cui, Sun, Shepard,
  Adam, Perona and Belongie}]{van2018inaturalist}
\bibinfo{author}{Van~Horn, G.}, \bibinfo{author}{Mac~Aodha, O.},
  \bibinfo{author}{Song, Y.}, \bibinfo{author}{Cui, Y.}, \bibinfo{author}{Sun,
  C.}, \bibinfo{author}{Shepard, A.}, \bibinfo{author}{Adam, H.},
  \bibinfo{author}{Perona, P.}, \bibinfo{author}{Belongie, S.},
  \bibinfo{year}{2018}.
\newblock \bibinfo{title}{The inaturalist species classification and detection
  dataset}, in: \bibinfo{booktitle}{Proceedings of the IEEE conference on
  computer vision and pattern recognition}, pp. \bibinfo{pages}{8769--8778}.
\bibitem[{Vaze et~al.(2022)Vaze, Han, Vedaldi and
  Zisserman}]{vaze2022generalized}
\bibinfo{author}{Vaze, S.}, \bibinfo{author}{Han, K.},
  \bibinfo{author}{Vedaldi, A.}, \bibinfo{author}{Zisserman, A.},
  \bibinfo{year}{2022}.
\newblock \bibinfo{title}{Generalized category discovery}, in:
  \bibinfo{booktitle}{Proceedings of the IEEE/CVF Conference on Computer Vision
  and Pattern Recognition}, pp. \bibinfo{pages}{7492--7501}.
\bibitem[{Wang et~al.(2021)Wang, Zhu, Zhao, Zeng, Wang and
  Tang}]{wang2021adaptive}
\bibinfo{author}{Wang, T.}, \bibinfo{author}{Zhu, Y.}, \bibinfo{author}{Zhao,
  C.}, \bibinfo{author}{Zeng, W.}, \bibinfo{author}{Wang, J.},
  \bibinfo{author}{Tang, M.}, \bibinfo{year}{2021}.
\newblock \bibinfo{title}{Adaptive class suppression loss for long-tail object
  detection}, in: \bibinfo{booktitle}{Proceedings of the IEEE/CVF conference on
  computer vision and pattern recognition}, pp. \bibinfo{pages}{3103--3112}.
\bibitem[{Wen et~al.(2023)Wen, Zhao and Qi}]{wen2023parametric}
\bibinfo{author}{Wen, X.}, \bibinfo{author}{Zhao, B.}, \bibinfo{author}{Qi,
  X.}, \bibinfo{year}{2023}.
\newblock \bibinfo{title}{Parametric classification for generalized category
  discovery: A baseline study}, in: \bibinfo{booktitle}{Proceedings of the
  IEEE/CVF International Conference on Computer Vision}, pp.
  \bibinfo{pages}{16590--16600}.
\bibitem[{Yang et~al.(2022)Yang, Zhu, Yu, Wu and Deng}]{yang2022divide}
\bibinfo{author}{Yang, M.}, \bibinfo{author}{Zhu, Y.}, \bibinfo{author}{Yu,
  J.}, \bibinfo{author}{Wu, A.}, \bibinfo{author}{Deng, C.},
  \bibinfo{year}{2022}.
\newblock \bibinfo{title}{Divide and conquer: Compositional experts for
  generalized novel class discovery}, in: \bibinfo{booktitle}{Proceedings of
  the IEEE/CVF Conference on Computer Vision and Pattern Recognition}, pp.
  \bibinfo{pages}{14268--14277}.
\bibitem[{Zhang et~al.(2023a)Zhang, Xu and He}]{zhangnovel}
\bibinfo{author}{Zhang, C.}, \bibinfo{author}{Xu, R.}, \bibinfo{author}{He,
  X.}, \bibinfo{year}{2023}a.
\newblock \bibinfo{title}{Novel class discovery for long-tailed recognition}.
\newblock \bibinfo{journal}{Transactions on Machine Learning Research} .
\bibitem[{Zhang et~al.(2023b)Zhang, Kang, Hooi, Yan and Feng}]{zhang2023deep}
\bibinfo{author}{Zhang, Y.}, \bibinfo{author}{Kang, B.}, \bibinfo{author}{Hooi,
  B.}, \bibinfo{author}{Yan, S.}, \bibinfo{author}{Feng, J.},
  \bibinfo{year}{2023}b.
\newblock \bibinfo{title}{Deep long-tailed learning: A survey}.
\newblock \bibinfo{journal}{IEEE transactions on pattern analysis and machine
  intelligence} \bibinfo{volume}{45}, \bibinfo{pages}{10795--10816}.
\bibitem[{Zhong et~al.(2021)Zhong, Cui, Liu and Jia}]{zhong2021improving}
\bibinfo{author}{Zhong, Z.}, \bibinfo{author}{Cui, J.}, \bibinfo{author}{Liu,
  S.}, \bibinfo{author}{Jia, J.}, \bibinfo{year}{2021}.
\newblock \bibinfo{title}{Improving calibration for long-tailed recognition},
  in: \bibinfo{booktitle}{Proceedings of the IEEE/CVF conference on computer
  vision and pattern recognition}, pp. \bibinfo{pages}{16489--16498}.
\bibitem[{Zhou et~al.(2020)Zhou, Cui, Wei and Chen}]{zhou2020bbn}
\bibinfo{author}{Zhou, B.}, \bibinfo{author}{Cui, Q.}, \bibinfo{author}{Wei,
  X.S.}, \bibinfo{author}{Chen, Z.M.}, \bibinfo{year}{2020}.
\newblock \bibinfo{title}{Bbn: Bilateral-branch network with cumulative
  learning for long-tailed visual recognition}, in:
  \bibinfo{booktitle}{Proceedings of the IEEE/CVF conference on computer vision
  and pattern recognition}, pp. \bibinfo{pages}{9719--9728}.
\bibitem[{Zhou et~al.(2017)Zhou, Lapedriza, Khosla, Oliva and
  Torralba}]{zhou2017places}
\bibinfo{author}{Zhou, B.}, \bibinfo{author}{Lapedriza, A.},
  \bibinfo{author}{Khosla, A.}, \bibinfo{author}{Oliva, A.},
  \bibinfo{author}{Torralba, A.}, \bibinfo{year}{2017}.
\newblock \bibinfo{title}{Places: A 10 million image database for scene
  recognition}.
\newblock \bibinfo{journal}{IEEE transactions on pattern analysis and machine
  intelligence} \bibinfo{volume}{40}, \bibinfo{pages}{1452--1464}.

\end{thebibliography}

\end{document}